\pdfoutput=1

\documentclass[11pt]{article}

\usepackage[]{acl}

\usepackage{times}
\usepackage{latexsym}

\usepackage[T1]{fontenc}

\usepackage[utf8]{inputenc}

\usepackage{microtype}

\usepackage[ruled,linesnumbered]{algorithm2e}
 
\usepackage{enumitem}
\usepackage{graphicx}
\usepackage[inkscapelatex=false]{svg}
\usepackage{multirow} 
\usepackage{subfigure}

\usepackage{bm}
\usepackage{amsmath,amsfonts}
\usepackage{booktabs}
\usepackage{xcolor}
\usepackage{amsfonts}
\usepackage{makecell}
\usepackage{mathrsfs}

\title{Spot the Difference: A Cooperative Object-Referring Game in Non-Perfectly Co-Observable Scene}

\author{Duo Zheng\textsuperscript{1}\thanks{ \ \ Work was done when Zheng was interning at Pattern Recognition Center, WeChat AI, Tencent Inc, China.}, Fandong Meng\textsuperscript{2}, Qingyi Si\textsuperscript{3}, Fairun Fan\textsuperscript{1}, \\
\bf{Zipeng Xu\textsuperscript{4}, Jie Zhou\textsuperscript{2}, Fangxiang Feng\textsuperscript{1}, Xiaojie Wang\textsuperscript{1}\thanks{ \ \ Xiaojie Wang is the corresponding author.}} \\
\textsuperscript{1}Beijing University of Posts and Telecommunications, Beijing, China \\
\textsuperscript{2}Pattern Recognition Center, WeChat AI, Tencent \\
\textsuperscript{3}Institute of Information Engineering, Chinese Academy of Sciences, Beijing, China \\
\textsuperscript{4}University of Trento \\
\texttt{\{zd, xjwang\}@bupt.edu.cn} \\
}

\begin{document}
\maketitle
\begin{abstract}
\vspace{-0.2cm}
Visual dialog has witnessed great progress after introducing various vision-oriented goals into the conversation, especially such as GuessWhich and GuessWhat, where the only image is visible by either and both of the questioner and the answerer, respectively. Researchers explore more on visual dialog tasks in such kind of single- or perfectly co-observable visual scene, while somewhat neglect the exploration on tasks of non-perfectly co-observable visual scene, where the images accessed by two agents may not be exactly the same, often occurred in practice. Although building common ground in non-perfectly co-observable visual scene through conversation is significant for advanced dialog agents, the lack of such dialog task and corresponding large-scale dataset makes it impossible to carry out in-depth research. To break this limitation, we propose an object-referring game in non-perfectly co-observable visual scene, where the goal is to spot the difference between the similar visual scenes through conversing in natural language. The task addresses challenges of the dialog strategy in non-perfectly co-observable visual scene and the ability of categorizing objects. Correspondingly, we construct a large-scale multimodal dataset, named \textit{SpotDiff}, which contains 87k Virtual Reality images and 97k dialogs generated by self-play. Finally, we give benchmark models for this task, and conduct extensive experiments to evaluate its performance as well as analyze its main challenges.
\end{abstract}

\section{Introduction}
Building a dialog agent that can intelligently communicate with people through comprehending and reasoning in vision and natural language is a challenging task in AI research \cite{DBLP:journals/corr/StrubVMPCP17, DBLP:journals/corr/abs-1812-02664}. Such visual dialog agents have broad prospects in social services and commercial applications, e.g., assisting the visually impaired people to understand the surroundings \cite{Bigham2010VizWizNR}, recommending products by dialog-based image retrieval \cite{guo2018dialogbased}, so that related researches \cite{Das2017VisualD, 2016GuessWhat, DBLP:journals/corr/abs-1902-00579, haber-etal-2019-photobook, ilinykh2019meetup, chen2020dmrm, DBLP:journals/corr/abs-2004-13278, DBLP:journals/corr/abs-2007-12750, takmaz2020refer, 2020DVD, liang2021maria, kottur2021simmc,chen-etal-2021-gog,xu2021modeling} have attracted increasing attention.


In recent years, researchers have proposed many visual dialog tasks for different scene settings, including single-observable scene and perfectly co-observable scene.
In single-observable scene, the scene is only visible to one interlocutor.
For example, \citet{Das2017VisualD} propose the task of Visual dialog, which requires the dialog agent to answer questions given an image and dialog history while the questioner can not see the image.
On the basis of the above task, GuessWhich \cite{Das2017LearningCV, murahari2019improving, zhou2019building, lee2019largescale, zheng-etal-2021-enhancing-visual} introduces an image-guessing game. This task aims at enabling the questioner imagine the invisible target image and finally guess it, through conversing with the answerer who could access the target image.
In co-observable scene, the scene is fully observed by all interlocutors.
For example, GuessWhat?! \cite{DBLP:journals/corr/abs-1711-07614,8639546,2017End,2018Beyond,2019What,10.1145/3394171.3413668} focuses on locating the target object in an image, which is visible by both the questioner and the answerer, through dialog between them.
\citet{DBLP:journals/corr/abs-2006-01460} introduce the task of SIMCC, which addresses the task-oriented dialog scenario on shopping domain where a system dialog agent and a user share the co-observable scene.


However, in actual applications, there are many situations where the visual scenes accessed by two people are similar but not be exactly the same.
Take the remote abnormal troubleshooting as an example, the user can access the problem machine, while the quality inspector can access intact machine. They determine the fault location through conversation online or by telephone.
At this time, it is more important to help each other to understand the partner's scene and clarify the differences, through dialog interaction.
Therefore, some researchers turn to investigate the visual dialog in such non-perfectly co-observable scene with the provision of a small-scale dataset.
\citet{lopes2018spot} study the dialog phenomenon under the setting of making two interlocutors to find differences between two similar scenes. They collect a dataset, which only contains 54 dialogs in 8 different cartoon scenes. More than that, lacking deeply analyzed challenges and corresponding solutions also makes its contribution to research community limited.

Two key challenges of the visual dialog in non-perfectly co-observable scene are not covered by the above tasks:
1) Difference-oriented dialog strategy. The two interlocutors participating in the dialog can only access their own part of the visual scenes, so they can only clarify the difference through the dialog. Therefore, to complete the goal of the dialog, the dialog interaction needs to constantly overcome the difficulty brought by differentiated visual information. 
2) Categorization-oriented question strategy. Human understands the world usually through categorization, which requires subjective generalization and classification of objects \cite{rosch1978cognition}. Such ability can be necessary for advanced agents. Therefore, finding a question strategy that can efficiently categorize the objects in the scene may be a critical path to quickly locating the difference. Although categorization has been mentioned in GuessWhat?!, all the questions in it are Yes/No questions, such as `is it a decoration?'. It ignores that an important purpose of categorization is induction, which often requires the abilities of accurate counting and clearly pointing out these objects, such as `I have three decorations, and you?', `what are they?'.
\begin{figure*}[] 
\centering 
\includegraphics[width=0.96\textwidth]{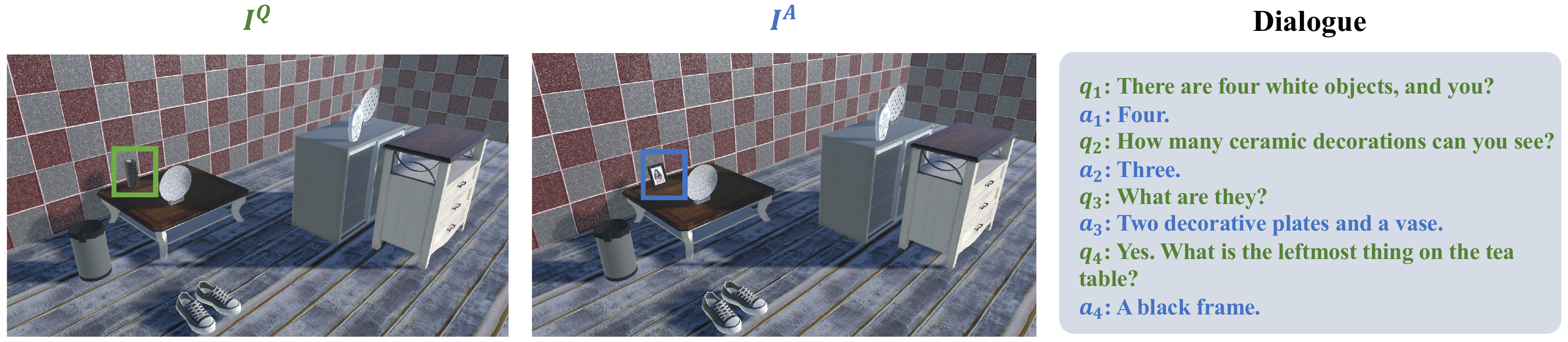} 
\caption{An example of \textit{Spot the Difference}. The green box indicates the different object from $I^Q$ to $I^A$. }
\label{dialog_instance}
\vspace{-0.4cm}
\end{figure*}

Obviously, the ability to deal with these challenges is significant for advanced dialog agents. To develop these capabilities of machines, in this paper, we propose an object-referring game -- \textit{Spot the Difference}. As shown in Figure \ref{dialog_instance}, the goal of \textit{Spot the Difference} is to spot the different object between two similar images via conversing in natural language  between a questioner and an answerer in a non-perfectly co-observable visual scene.
To this end, we construct a large-scale multimodal dataset, named \textit{SpotDiff}, which contains 87k images and 78k dialogs. 
First, we generate the images of \textit{SpotDiff} with an elaborately designed scene simulator, taking into account the coherence of the real world.
Then, based on the generated images, we generate the dialogs of \textit{SpotDiff} through a well-designed two-stage dialog generation algorithm.
Finally, we propose benchmark models for \textit{Spot the Difference}, which are based on the multimodal pre-trained model LXMERT \cite{tan2019lxmert}. We evaluate the performance of the dialog system and the answerer agent, and analyze the model's ability in dialog strategy and categorization.

\begin{figure*}[] 
\centering 
\includegraphics[width=0.97\textwidth]{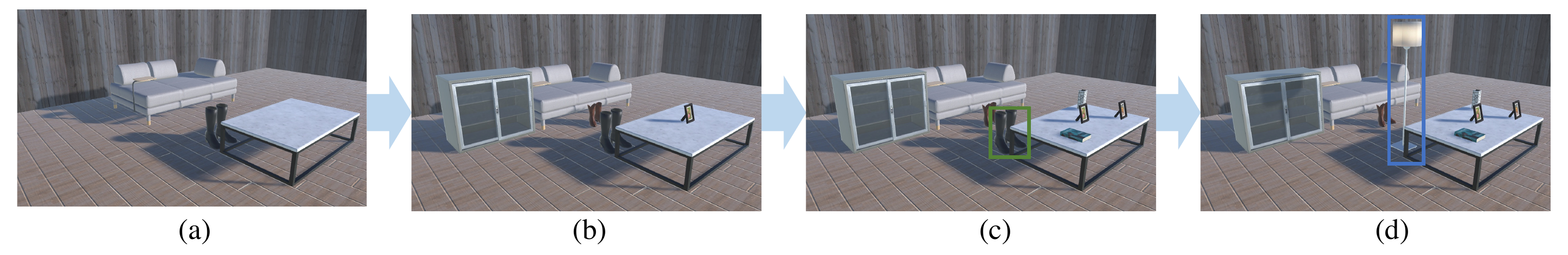} 
\caption{The generation process of \textit{SpotDiff} images. (a)--(c) show the object-by-object generation process of scene. (c) and (d) are a pair of similar images. The green box in (c) represents the different object from (c) to (d).}
\label{image_gen}
\vspace{-0.4cm}
\end{figure*}

Our main contributions are concluded as follows\footnote{The code and data are publicly available at: \url{https://github.com/zd11024/Spot_Difference}}:
\begin{itemize}[leftmargin=*]
\vspace{-0.2cm}
\setlength{\itemsep}{0pt}
\setlength{\parsep}{0pt}
\setlength{\parskip}{0pt}
\item We propose a new visual dialog task -- \textit{Spot the Difference}, which mainly addresses challenges of the dialog strategy in non-perfectly co-observable visual scene and the ability of categorizing objects.
\item We construct the \textit{SpotDiff} dataset, which consists of 87k Virtual Reality images and 95k programmatically simulated dialogs.
\item We provide strong benchmark models for  \textit{Spot the Difference}. 
Experimental results show that the task performance can be improved by designing  difference-oriented dialog strategy and categorization-oriented question strategy, both of which are the challenges that the task of non-perfectly co-observable scene hope to address. These provide insights for developing more intelligent visual dialog agents.
\end{itemize}

\section{\textit{Spot the Difference} Game}
As illustrated in Figure \ref{dialog_instance}, \textit{Spot the Difference} is an object-referring game conducted by a questioner and an answerer.
The questioner and answerer can see images $I^Q$ and $I^A$, respectively.

The goal of questioner is to spot the difference from $I^Q$ to $I^A$, i.e., the object in $I^Q$ that is not in $I^A$ (marked with a green box in Figure \ref{dialog_instance}). The questioner constantly asks questions based on the image $I^Q$, such as asking the number of objects with specific conditions, the referential content of the previous round, and the object at a specific location, e.g., $q_1$ -- `There are four white objects?', $q_3$ -- `What are they?', $q_4$ -- `What is the leftmost thing on the tea table?'. After the questioner has located the different object, it terminates the conversation and makes a guess on the correct object list of $I^Q$.

Based on the image $I^A$ and the question, the answerer gives the answer, which may be a number, a description of one or multiple objects, e.g., $a_1$ -- `Four', $a_3$ -- `Two decorative plates and a vase'.


\section{\textit{SpotDiff} Dataset}
In this section, we first describe how images and dialogs of the \textit{SpotDiff} are generated in Section \ref{sec_image_generation} and Section \ref{sec_dialog_generation}, respectively. Then we present the dataset analysis in Section \ref{sec_dataset_analysis}.
\subsection{Image Generation} \label{sec_image_generation}

First, we develop a scene simulator to generate \textit{SpotDiff} images in Virtual Reality (VR) environment. Then for each image, we construct its scene graph, serving as the input to dialog generation.

\subsubsection{Scene Simulator}
\label{scene_simulator}
The scene simulator generates similar image pairs with only one object different and the generation process of \textit{SpotDiff} images is illustrated in Figure \ref{image_gen}.
First, the scene simulator generates a random scene by placing objects item by item.
Then, it randomly selects one object from all the objects that can be replaced in the scene, and replace it with a random object of a different category, or that of the same category with different attributes.
Finally, the scene simulator renders the scene in Unity3D \cite{unity} and takes screenshots with Unity Perception\footnote{A toolkit provided by Unity Corporation for generating large-scale computer vision datasets.} \cite{com.unity.perception2021}.

Real world scenes appear as a composite of coherent objects \cite{4587799}. To make VR scenes more reality, the scene simulator adopts an elaborately designed search algorithm, mainly considering the following aspects:


\noindent \textbf{Richness of Objects.} To generate richer scenes, more diverse objects are needed. We use 251 digital assets\footnote{We obtain them from \url{https://assetstore.unity.com/} and \url{https://www.turbosquid.com/}} which belong to 86 different categories.
\noindent \textbf{Placement Relationship.} A directed graph (please refer to Appendix \ref{placement_rel} for details) is defined to describe the placement relationship between categories. For example, bread can be placed on a plate, but not directly on the floor. 
\begin{figure*}[] 
\centering 
\includegraphics[width=0.95\textwidth]{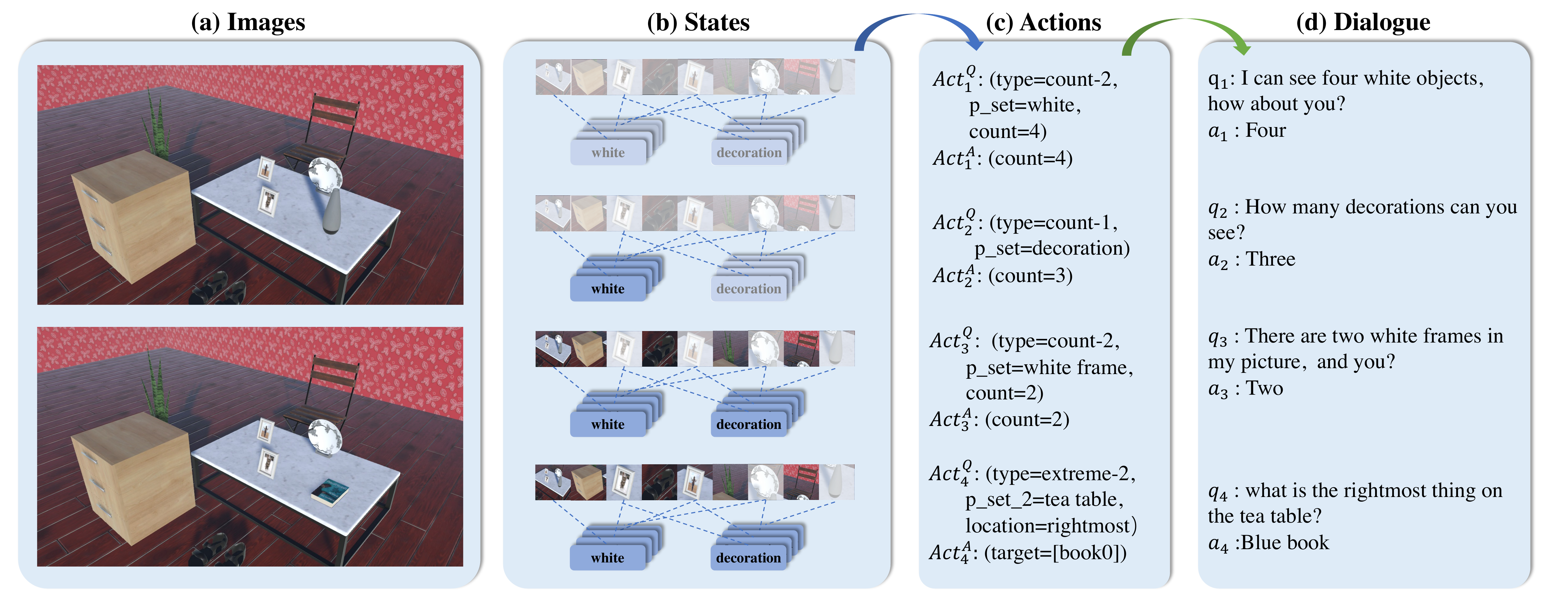} 
\caption{The generation process of a \textit{SpotDiff} dialog. (a) The questioner simulator and answerer simulator look at the top and bottom images, respectively. (b) The states give part of the visual state tracking information under current dialog, where the transparent node means that an object or property has not been fully confirmed. (c) $Act_t^Q$ and $Act_t^A$ are respectively question action and answer action generated at time $t$. (d) The dialog consists of a series of question-answer pairs, where the question $q_t$ and answer $a_t$ are mapped from $Act_t^Q$ and $Act_t^A$, respectively.}
\label{dialog_gen}
\vspace{-0.5cm}
\end{figure*}

\noindent \textbf{Spatial Arrangement\footnote{We present the implementation details of spatial arrangement and object co-occurrence in Appendix \ref{spatial_arrangement} and \ref{co-occur}, respectively. \label{foot2}}.} The scene should neither be too evacuated nor too compact. The former may cause the pixels of objects in the image to be too small, and the latter may cause mutual occlusion between objects.

\noindent \textbf{Object Co-Occurrence\textsuperscript{\ref{foot2}}.}  Related objects co-occur with high probability. For example, computers, mice and keyboards often appear together because they are all office supplies.

\subsubsection{Image Scene Graph}
\label{scene_graph_case}
The scene graph\footnote{We show an example of scene graph in Appendix \ref{img_app}.} contains the information of all objects in an image, including:

\noindent 1) Attribute: Each Object is annotated with color and material.

\noindent 2) Taxonomy: Taxonomy information is depicted by a predefined hierarchical tree of categories, which is in Appendix  \ref{object_category}. For example, pizza belongs to $\{\textit{pizza, baked food, food} \}$.

\noindent 3) Position: Position information is described by 2D bounding box and 3D bounding box, which are annotated when generating images with Unity Perception \cite{com.unity.perception2021}.

Color, material and categories are regarded as atomic properties of an object.

\begin{figure*}[] 
\centering 
\includegraphics[width=0.95\textwidth]{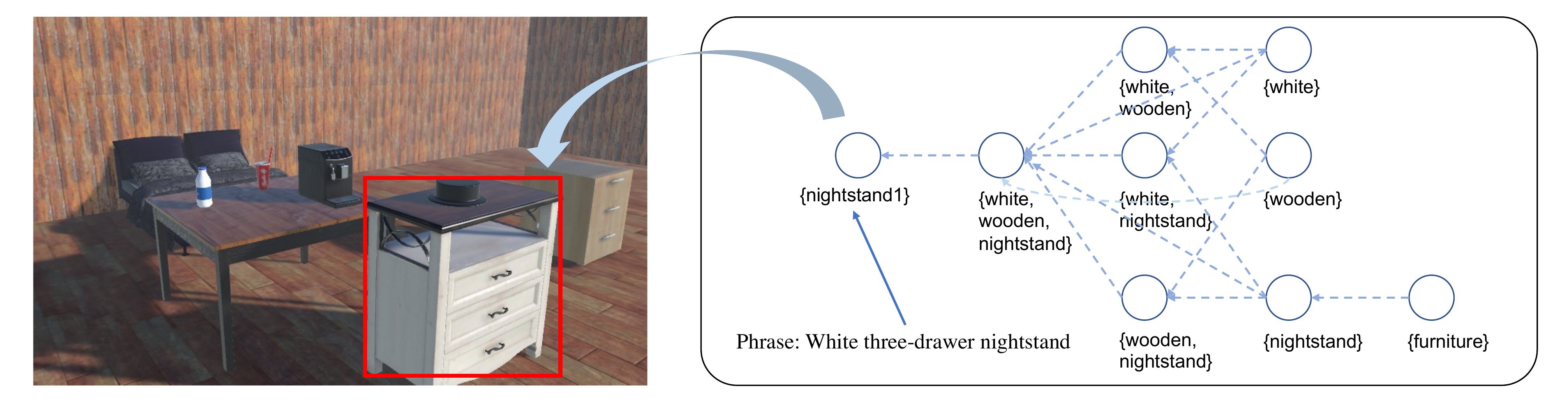} 
\caption{A simplified instance of object state graph.}
\label{graph}
\vspace{-0.2cm}
\end{figure*}

\begin{figure*}[] 
\centering 
\includegraphics[width=0.9\textwidth]{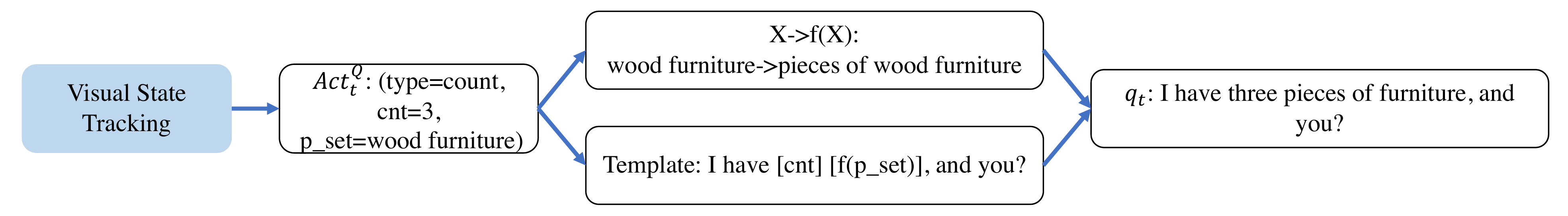} 
\caption{The process of transforming a question action to a natural language sentence. }
\label{nlg}
\vspace{-0.5cm}
\end{figure*}

\subsection{dialog Generation} \label{sec_dialog_generation}
With the image scene graph as input, we design a two-stage dialog generation approach as shown in Figure \ref{dialog_gen}. In the first stage, a questioner simulator and an answerer simulator are used to generate a dialog action sequence through self-play (Section \ref{dialog_action_generation}). In the second stage, the dialog action sequence is mapped to natural language through manually defined templates (Section \ref{natural_language_generation}). 

\subsubsection{Dialog Action Generation} \label{dialog_action_generation}
Inspired by previous works \cite{moon2020situated, kottur2021simmc}, the dialog action sequence consists of question actions and answer actions, both of which are composed of a series of slot-value pairs.
The dialog action sequence is interactively generated by the questioner simulator and answerer simulator.
In concrete, at each round, the questioner simulator produces a question action and the answerer simulator gives the corresponding answer action.
The interaction is repeated until the questioner simulator could locate the target object.

Question actions are divided into nine subtypes, which belong to four types:
1) \textbf{Count} type (count-nohint and -hint) asks the number of objects with specific properties. Comparing with count-nohint type (e.g., `how many white objects can you see?'), count-hint type adds a hint for counting, e.g., `I have four white objects, how about you?'.
2) \textbf{Extreme} type (extreme-pic, -obj and -obj2) asks for a specific description of the object at a positional extreme among conditioned objects. For extreme-pic type, the conditioned objects are all objects in the picture, e.g., `what is the rightmost thing?'.
For extreme-2 and -3 types, the conditioned objects are objects placed on a given object, e.g., `what is the rightmost thing on the tea table?'.
3) \textbf{Query} type (query-color and -material) asks for the color or attribute of the referent, which may not exist or cannot be uniquely determined.
4) \textbf{Refer} type (ref-it and -them) follows the count type, and requires the answerer to give a specific description of the objects referred to in the previous round. Ref-it type asks one object while ref-them asks multiple, e.g., `what is it?' and `what are they?'.

At each round, the questioner simulator tracks visual state according to dialog history, then selects an appropriate question action based on the tracked visual state and question strategy. A good questioner simulator can achieve the above goal by answering the following questions. $\mathbf{Q_1}$: How to accurately track the state of each object in an image, taking into account entailment relationships between properties of the object. For example, one won't ask `is there any fruit?' after knowing there is an apple;  $\mathbf{Q_2}$: How to efficiently guide the conversation to avoid object-by-object mechanical enumeration.

\noindent \textbf{$\mathbf{Q_1}$: Visual State Tracking.}
To maintain the state of the image as the dialog proceeds, the questioner simulator constructs an object state graph for each object.
We define a property set as a combination of properties or an identifier for the object itself.
As shown in Figure \ref{graph}, the nightstand in red contain many property set, e.g, \{white, nightstand\} or \{nightstand1\}.
Obviously, there are entailment relationships between property sets, inspiring us to describe the state of an object with a directed graph in the process of dialog.

To this end, we construct the object state graph for each object, where nodes represent property sets and edges represent entailment relationships between them.
To clarify which property sets of an object are known or not, each node maintains a boolean value initialized to False, which we name as confirmation status.
When a node is confirmed (its confirmation status is True), all reachable nodes from it are also confirmed.
Conversely, for two confirmed nodes, whose property sets are denoted as $A$ and $B$ respectively, the node corresponding to the property set $\vert A \cup B \vert$ is also confirmed. In addition, each node corresponds to a description template set, which is used to map the property set to a phrase in the natural language generation stage, such as \{wooden,furniture\} $\rightarrow$ \textit{pieces of wooden furniture }.



In addition, the questioner simulator also track a candidate object set $S_{cand}$, which includes objects whose existence has not been confirmed. In the beginning, $S_{cand}$ includes all objects in the image. As the dialog proceeds, an object will be removed from $S_{cand}$ after all nodes in its corresponding object state graph have been confirmed.




\noindent \textbf{$\mathbf{Q_2}$: Categorization-Based Question Strategy.}
For efficient questioning, we propose a categorization-based question strategy whose main idea is to gather more information by generalizing half of the remaining objects as much as possible. As illustrated in Figure \ref{dialog_gen}, $q_2$ -- `How many decorations can you see' generalizes the decorations in the image to confirm whether a decoration has been replaced.
Therefore, we design an approach to simulate such a strategy.
In concrete, the questioner simulator maintains a list of question types that could be performed. The count type is always in the list. 
When the size of the candidate object set $S_{cand}$ is less than $n$, the extreme and refer type is added to the list\textsuperscript{\ref{footnote_hyper_parameter}.} 
When the question type of the previous round is count and the corresponding answer is less than $m$, the refer type is added to the list\footnote{$n$ and $m$ are hyper-parameters, which are empirically set to 5 and 4, respectively. \label{footnote_hyper_parameter}.}
The final question type is sampled from the list.
After the question type is determined, the slot-value pairs are heuristically obtained as follows:
\begin{itemize}[leftmargin=*]
\vspace{-0.2cm}
\setlength{\itemsep}{0pt}
\setlength{\parsep}{0pt}
\setlength{\parskip}{0pt}
\item Count type: First, the questioner simulator counts frequencies of all property sets, which are defined as the number of unconfirmed nodes corresponding to the property set for objects in the candidate object set $S_{cand}$.
Then, it chooses the property set whose frequency is closet to $\frac{\vert S_{cand} \vert}{2}$, to produce a question action.
\item Extreme/Query/Refer type: First, the questioner simulator maintains a candidate list to store all valid slot-value pairs.
Then, it enumerates all slot-value pairs, and put slot-value pairs that could collect new information into the list.
The final slot-value pairs are sampled from the list.
\vspace{-0.2cm}
\end{itemize}

Given a question action, the answerer simulator retrieves the corresponding information on the image scene graph, and produce the answer action (Please refer to Appendix \ref{answer_action} for details).

\subsubsection{Natural Language Generation} \label{natural_language_generation} 
At this stage, each action is mapped to a natural language sentence. 
Taking question action as an example (see Figure \ref{nlg}), we randomly select a question template according to the question subtype and fill the slot values into the question template to produce a question. 
Notably, the property sets are mapped to natural language phrases with description template sets.
In addition, to make dialogs more fluid, we design transition sentences to concatenate adjacent rounds of dialog. There are 43 templates and 1659 human-annotated natural language phrases in total. We refer readers to Appendix \ref{all_templates} for the details of templates.


\subsection{Dataset Analysis} \label{sec_dataset_analysis}
For each \textit{SpotDiff} image pair, we generate 2 dialogs by changing the order between images, and the dialogs that fail to complete the task within 10 rounds are discarded. The \textit{SpotDiff} dataset contains 78k dialogs and 87k \textit{SpotDiff} images, and is splited by randomly assigning 80\%, 10\% and 10\% of image pairs and its corresponding dialogs to train, valid and test set.
Table 2 shows the comparison results of \textit{SpotDiff} with SIMCC 2.0 \cite{kottur2021simmc} and CLEVR dialog \cite{kottur2019clevrdialog}. 
The SpotDiff dataset has much more unique answers than CLEVR Dialog (4.0k vs 29), indicating the answerer in our task has a higher degree of freedom.

\begin{table}[]
\renewcommand\arraystretch{1.}
\centering
\resizebox{0.4\textwidth}{!}{
\setlength{\tabcolsep}{1.3mm}{
    \begin{tabular}{l|ccc} \hline
        & \textbf{\textit{SpotDiff}} & \textbf{SIMCC2} & \textbf{CLEVR} \\ \hline
        \# dialogs & 78k & 11k & 425k \\
        \# Images & 87k & 1.5k & 85k \\
        \# Turn & 5.36 & 5.2 & 10 \\
        \# Unique Q & 32k & - & 73k \\
        \# Unique A & 4.0k & - & 29 \\
        Avg \# Q len & 9.8 & - & 10.6 \\  \hline
    \end{tabular}
}
}
\caption{Comparison of \textit{SpotDiff} to similar datasets.} 
\end{table}

\begin{figure}[] 
\centering 
\includegraphics[width=0.41\textwidth]{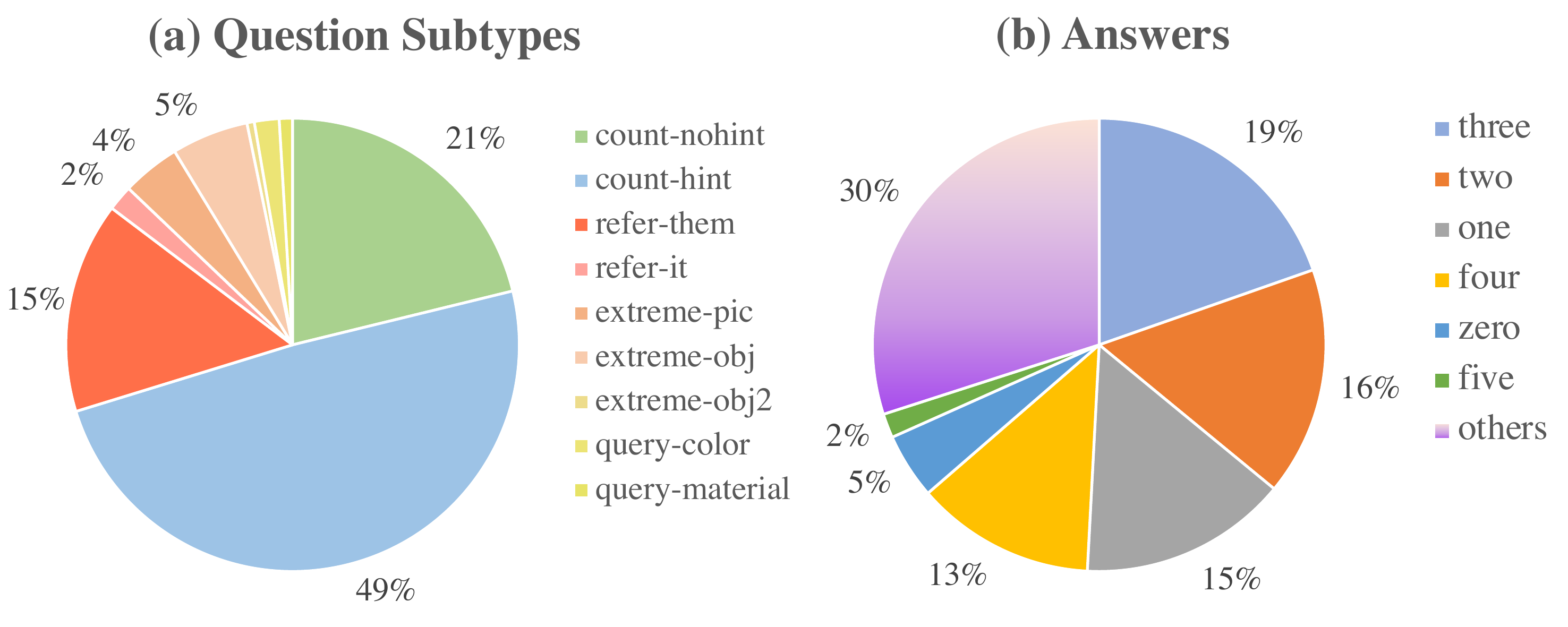} 
\caption{Distribution of question subtypes and answers.}
\label{data_stat}
\vspace{-0.4cm}
\end{figure}

Figure \ref{data_stat} (a) shows the distribution on question subtypes. More than 70\% of the questions in the \textit{SpotDiff} dataset need to count objects with specific properties. Figure \ref{data_stat} 
(b) presents the distribution of answers. 
There are a total of 4.0k unique answers, of which the 6 most frequent unique answer account for 69\% of the total answers, while remaining unique answers account for 30\%, making the distribution a long-tailed distribution. 

\section{Task Formulation}



Following previous work \cite{2016GuessWhat}, the Questioner Bot (Q-Bot) consists of Question Generator (QGen) and Guesser, which are responsible for asking questions and guessing the target object, respectively. The Answerer Bot (A-Bot) is a Visual Question Answering (VQA) model.

\noindent \textbf{QGen.}
At round $t$, QGen asks a question $q_t$ given the dialog history $H_{t-1}=\{(q_1,a_1), \cdots, (q_{t-1}, a_{t-1})\}$ and the image $I^Q$, which could be formulated as:
\begin{equation}
q_t \sim P(q|H_{t-1}, I^Q).
\end{equation}
\noindent \textbf{A-Bot.}
A-Bot predicts the answer $a_t$ from the candidate answer set, based on the question $q_t$, dialog history $H_{t-1}$, and the image $I^A$, which could be denoted as:
\begin{equation}
a_t \sim P(a|q_t, H_{t-1}, I^A).
\end{equation}

\noindent \textbf{Guesser.}
After $T$ rounds of dialog, Guesser makes a guess on the correct object list $O_{correct}$ of $I^Q$ given the full dialog history $H_T$ as follow:
\begin{equation} \label{guesser_eq}
o^* \sim P(o|H_T, O_{correct}),
\end{equation}
where $T$ is the maximum number of dialog rounds, $O_{correct}=\{ (c_1, p_1),\cdots, (c_M, p_M) \}$, $c_i$ and $p_i$ are the correct category and relative bounding box of the $i$-th object, respectively.

\section{Experiments}
To explore the challenges arising from the task, we first train benchmark models and evaluate their performance on \textit{SpotDiff} dataset. Then we conduct extensive experiments to analyze two main challenges: categorization and dialog strategy.

\subsection{Benchmark Models}
We train benchmark models on \textit{SpotDiff} dataset:
1) \textbf{QGen}\textsuperscript{\ref{foot3}}: 
A LXMERT \cite{tan2019lxmert}-initialized encoder paired with a randomly initialized decoder. We also try initializing the decoder with pre-trained models, such as BERT \cite{devlin2019bert} and GPT-2 \cite{radford2019language}, but no gain is observed.
2) \textbf{A-Bot}\footnote{We also adopts GPT-2 \cite{radford2019language} and UpDn \cite{anderson2018bottomup} as QGen and A-Bot, respectively. Please refer to Appendix \ref{abot_app} and \ref{dialog_system_comp} for details. \label{foot3}}: A VQA model with the multimodal pre-trained model LXMERT as encoder and an MLP head to predict the answer.
3) \textbf{Guesser}: A BERT \cite{devlin2019bert} encoder with a classification head to predict the target object. The three components are trained separately, please refer \ref{implementation_details} to for implementation details.


\begin{table}[]
\centering
\renewcommand\arraystretch{1}
\setlength{\tabcolsep}{3.4mm}{
\begin{tabular}{l|ccc|c} \hline
\# & GT-Q & GT-A & GT-V & SUCC $\uparrow$ \\ \hline
1 & - & - & - & 33.70 \\ 
2 & - & - & $\surd$ & 42.01 \\
3 & $\surd$ & - & - & 63.16 \\
4 & - & $\surd$ & - & 32.67 \\
5 & $\surd$ & $\surd$ & - & 71.26 \\ \hline
\end{tabular}}
\caption{The performance of the dialog system\protect\footnotemark.
GT-Q: ground truth question, GT-A: ground truth answer, GT-V: visual features extracted by ground truth box, SUCC: task success rate (\%). 
$\uparrow$: higher is better.
}
\label{dialog_system_performance}
\vspace{-0.4cm}
\end{table}

\begin{figure}[] 
\centering 
\includegraphics[width=.47\textwidth]{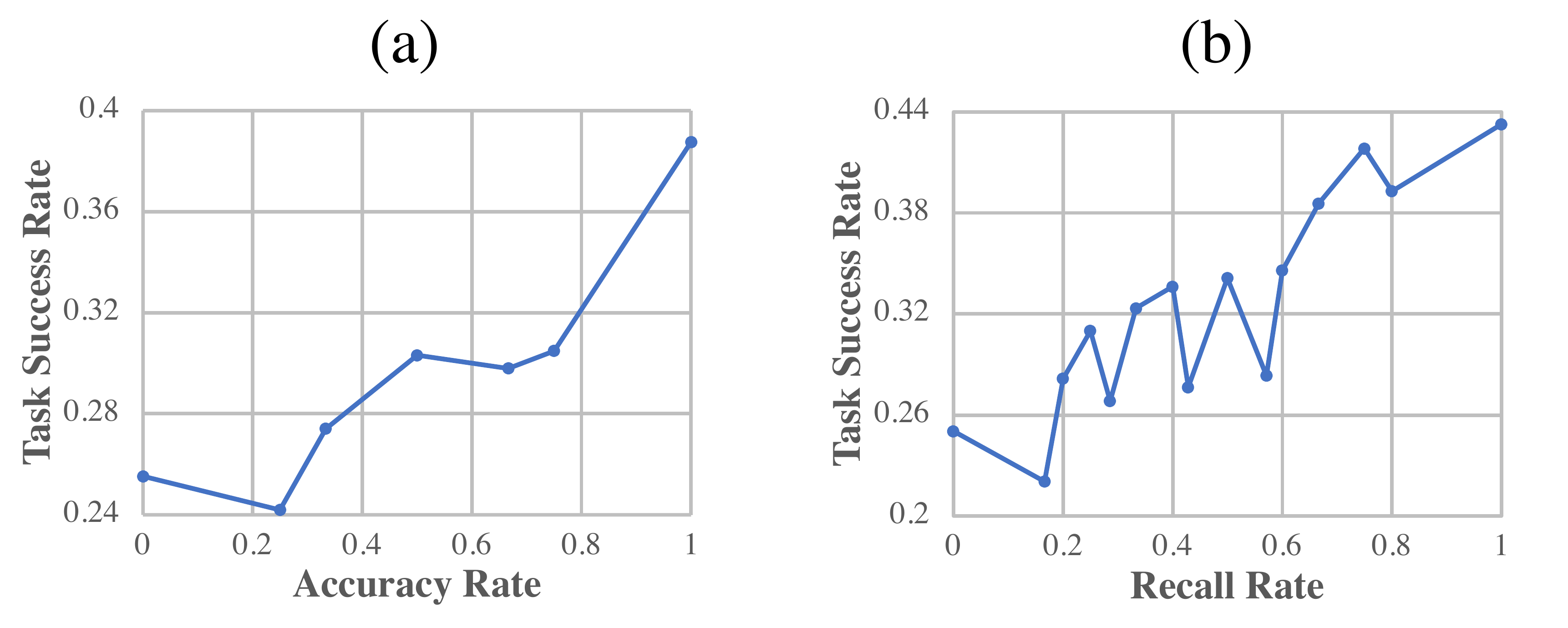} 
\caption{(a)/(b) shows the relationship between the task success rate and the accuracy/recall of Cate-Q.}
\label{categorization}
\vspace{-0.4cm}
\end{figure}


\begin{figure*}[] 
\centering 
\includegraphics[width=0.93\textwidth]{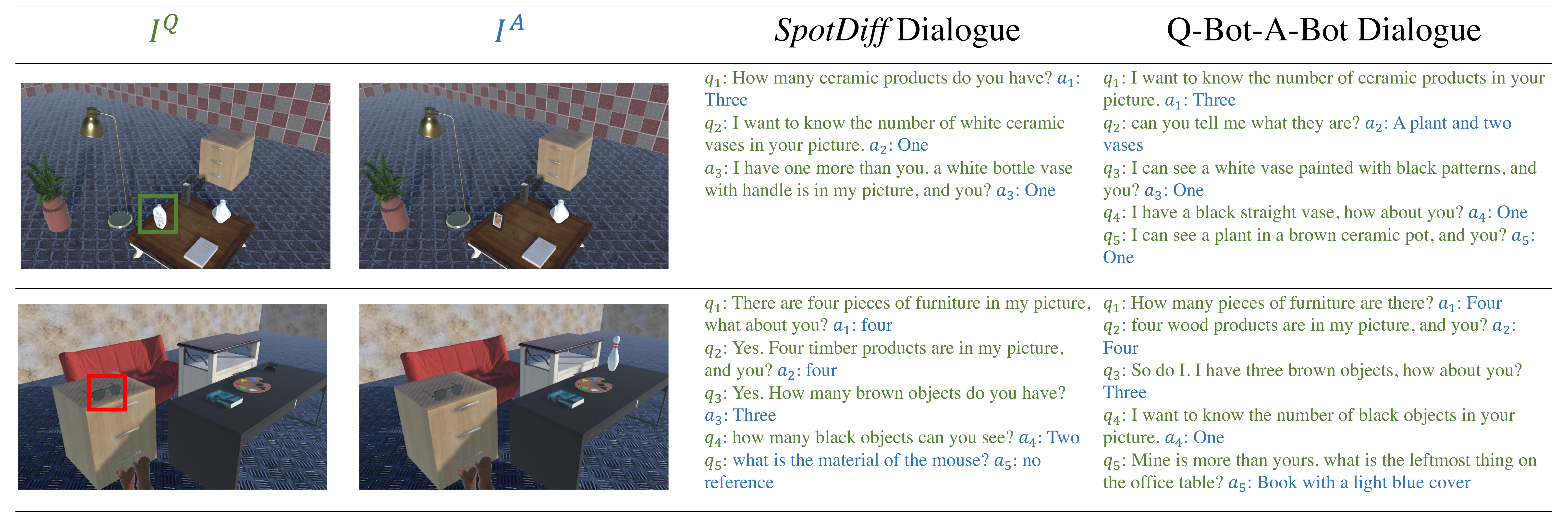} 
\caption{Examples of \textit{SpotDiff} dialogs and Q-Bot-A-Bot dialogs. Green sentences: questions, blue sentences: answers, red box: the wrong guess, green box: the correct guess.}
\label{main_case_study}
\vspace{-0.4cm}
\end{figure*}


\subsection{Dialog System Performance.}
We investigate the performance of the dialog system under the setting of \textit{Spot the Difference}. Specifically, QGen and A-Bot first interactively generate a 5-round Q-Bot-A-Bot dialog, and then Guesser makes a guess on the correct object list given the generated dialog.
Table \ref{dialog_system_performance} shows the task success rate under different settings.
GT-Q and GT-A indicate whether the ground truth question and ground truth answer are used, respectively. 
GT-V indicates whether the visual features are extracted by the ground truth bounding box.
Comparing row 1 and row 2, it can be seen that correct object detection could improve task success rate. 
Comparing row 1/4 and row 3, it shows that the questioner model greatly limits the task success rate and the main challenge of the task lies in the modeling of QGen. 
Comparing row 1 and 5, there is still a large gap between the Q-Bot-A-Bot dialog and the ground truth data.

Counterintuitively, the GT-A\footnote{The generated answers are corrected by the answerer simulator.} variant underperforms the baseline, which may be caused by the inconsistency between the wrong question and the correct answer. For example, for a Q-Bot-generated question-`I have three decorations, and you?', the A-Bot would select answer from \{two, three, four\}, but the actual answer could be \{zero, one\}.

\subsection{A-Bot Performance.}
Although the challenges of our task lie in the modeling of the questioner, we verify A-Bot performance in term of accuracy on various question subtypes under classification setting. We observe that: 1) Count-hint (87.59\%) surpasses count-nohint (83.23\%) due to the hints about counting in count-nohint questions. 2) Extreme-pic (82.92\%) outperforms extreme-obj (81.54\%) and extreme-obj2 (76.75\%) because the model's spatial reasoning ability is more urgently required for extreme-obj and -obj2. 3) Query-color and -material achieve the accuracy of 91.02\% and 88.54\%, respectively. 4) Refer-it (89.78\%) is better than refer-them (85.63\%), considering that refer-them questions ask multiple objects while refer-it questions ask one.

\subsection{Effect of Categorization}
We name the \textit{count} question that could generalize at least two different kinds of objects on an image as a Cate-Q. 
To investigate the effect of categorization ability on task success rate, we first obtain Q-Bot-A-Bot dialogs on the test set, and then group these dialogs in different ways, i.e., the accuracy rate and recall rate of Cate-Q in the dialog.

\noindent \textbf{Accuracy Rate of Cate-Q.}
For each Q-Bot-A-Bot dialog, we first extract quantifiers and property sets in all Cate-Q, and calculate the accuracy over Cate-Q in a dialog by matching with objects on the scene graph.
Then we divide the Q-Bot-A-Bot dialogs according to the counting accuracy of Cate-Q, and examine the task success rate of each group. As shown in Figure \ref{categorization} (a), as the accuracy of Cate-Q increases, the task success rate shows an increasing trend, demonstrating that accurate counting for Cate-Q could help to complete the task.

\noindent \textbf{Recall Rate of Cate-Q.} 
We extract the property sets of the Cate-Q for each Q-Bot-A-Bot dialog and corresponding ground truth dialog, which are denoted as $A$ and $B$, respectively. We define the recall rate of Cate-Q as $\frac{\vert A \cap B \vert}{\vert B \vert}$. The Figure \ref{categorization} (b) shows the task success rate increases as the the recall rate increases, indicating the importance of selecting appropriate property sets to raise Cate-Q for successfully completing the task.


\subsection{Effect of Dialog Strategy}
\noindent \textbf{Action Transition.}
To investigate the relationship between action transition and task success rate, we group Q-Bot-A-Bot dialogs according to adjacent question action transitions. Question action transitions could be divided into deepening action transitions and converting action transitions according to whether the latter question deepens the previous one. 
Table \ref{action_transition} shows that dialogs with deepening action transitions achieve higher task success rate (35.85\% vs 32.72\%) because the deepening action transitions could help Q-Bot to narrow the scope of the target object.

\begin{table}[]
\small
\renewcommand\arraystretch{1}
\centering
\setlength{\tabcolsep}{2.5mm}{
    \begin{tabular}{lll|c} \hline
    \multicolumn{3}{c|}{Action Transition} & SUCC$\uparrow$ \\ \hline
    furniture & $\rightarrow$ & brown wooden furniture & 40.75 \\
    black & $\rightarrow$ & black decoration & 39.70 \\ 
    \multicolumn{3}{c|}{$\dots$} & \\ \hline
    \multicolumn{3}{c|}{deepening action transition} & 35.82 \\ \hline\hline
    white & $\rightarrow$ & furniture & 32.43 \\ 
    furniture & $\rightarrow$ & toy & 29.38 \\ 
    \multicolumn{3}{c|}{$\dots$} & \\ \hline
    \multicolumn{3}{c|}{converting action transition} & 32.72 \\ \hline
    \end{tabular}
}
\caption{The relationship between action transition and task success rate. SUCC: task success rate (\%).} 
\label{action_transition}
\end{table}

\noindent \textbf{Case Study.}
We conduct case studies to investigate the effect of dialog strategies in Figure~\ref{main_case_study}. In the first Q-Bot-A-Bot dialog, the questioner successfully complete the task by gradually narrowing down the candidates. In the second Q-Bot-A-Bot dialog, $q_5$ asks about the green book instead of the black mouse after finding the difference in black objects.

\section{Related Work}
\noindent \textbf{VQA and Captioning.} Many work have been introduced for studying vision-and-language understanding, including VQA \cite{DBLP:journals/corr/GoyalKSBP16, anderson2018bottomup, 2019GQA} and image captioning \cite{2014Show, tan2019expressing, li2020contextaware}. \citet{DBLP:journals/corr/abs-1808-10584} propose the task to describe the difference between two similar images and collect the spotting-the-difference dataset.
In contrast, under our setting, the questioner could only access one image and understand the content of another image through dialog interaction.

\noindent \textbf{Collaborative Dialog in Visual Scene.} Numerical work \cite{2016GuessWhat, Das2017VisualD} pay attention on dialog in single or partially co-observable visual scene. Recently, some researchers focus on dialog in non-perfectly co-observable scene. \citet{haber-etal-2019-photobook} introduce the PhotoBook dataset, whose goal is to determine the shared images through conversation between two interlocutors. \citet{ilinykh2019meetup} propose a two-player coordination game, named MeetUp!, and collect a multimodal corpus that contains 430 dialogs. The player can switch the unshared scene by moving in a virtual environment, and meet with each other in a specific location during interaction.

\noindent \textbf{Collaborative Dialog in Abstract Context.} Our work is also related to partially-observable collaborative game in abstract context \cite{HeHe2017LearningSC, 2019A,TakumaUdagawa2020AnAC, fried2021referencecentric}. \citet{2019A} introduce ONECOMMON, which addresses the challenges of dialog technology in continuous and partially-observable context. In this task, two players have different views of a game board, which consists of multiple dots described in continuous value.

\section{Conclusion}
In this paper, we propose a cooperative object-referring game -- \textit{Spot the Difference}, where the goal is to locate the different object between two similar images via conversing between questioner and answerer. The task addresses two challenges at visual dialog in non-perfectly co-observable scene, including the difference-oriented dialog strategy and the ability of categorization. We construct a large-scale dialog dataset \textit{SpotDiff}, which contains 87k images and 78k dialogs. Additionally, we provide strong benchmark models and conduct extensive experiments to analyze the two key challenges.

\bibliography{anthology,custom}
\bibliographystyle{acl_natbib}

\newpage
\appendix

\section{Image Generation}

\subsection{Taxonomy Information} \label{object_category}
We present the taxonomy information with a predefined hierarchical tree structure, which is illustrated in Table \ref{table_taxonomy}.

\subsection{Placement Relationship} \label{placement_rel}
We empirically construct a directed graph of object placement relationship. As shown in Table \ref{table_placement}, it describes which category of objects could be placed on a object of specific category.

\subsection{Spatial Arrangement} \label{spatial_arrangement}
Given an object $o$ (the object to be placed), a rectangular area and existing objects, we put the unplaced object on the area as follows:
\begin{itemize}
\setlength{\itemsep}{0pt}
\setlength{\parsep}{0pt}
\setlength{\parskip}{0pt}
\item[1)] Randomly sample $T$ points on the area.
\item[2)] Filter out the points whose distance from any existing object is less than X or that cross the boundary.
\item[3)] Select the point with the minimum L1 distance to its closet existing object, and place the object $o$ at the point.
\end{itemize}

After taking screenshots, we retain the image where the pixels of objects in the image are all larger than $Y$, to avoid the serious mutual occlusion between objects in the image.

\subsection{Object Co-Occurrence} \label{co-occur}
Considering the hierarchical tree structure of categories, we define the degree of divergence $d_u$ for category $u$ as:
\begin{equation}
    d_u = \sum_{v \in child(u)} [\exists o \in O \wedge v \in f(o)],
\end{equation}
where $child(u)$ means the child categories of the category $u$ (e.g., $child(fruit)=\{apple, banana\}$), $O$ is the object list of the image, $f(o)$ is the category set corresponding the object $o$ (e.g., for an apple, its category set is $\{apple, fruit  , food\}$), $[\cdot]$ is 1 if and only the expression in the bracket is True.

To make the related objects co-occur with high probability, for each category $u$, we limit $d_u$ not to exceed K=3.

\begin{figure*}[] 
\centering 
\includegraphics[width=1.\textwidth]{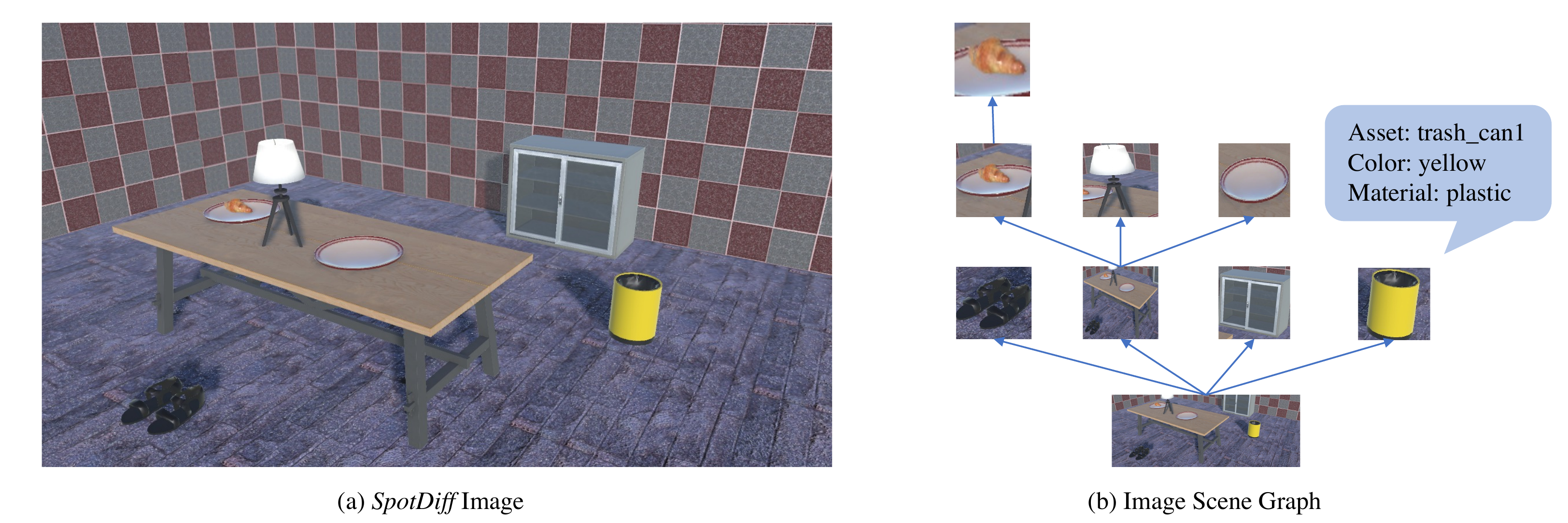} 
\caption{An Example of image scene graph. (a) gives a \textit{SpotDiff} image and (b) displays its corresponding scene graph, where blue lines indicate placement relationships between objects.}
\label{scene_graph}
\end{figure*}

\begin{table}[]
\centering
\renewcommand\arraystretch{1}
\setlength{\tabcolsep}{3.5mm}{
\begin{tabular}{l|cc|c} \hline
\# &QGen & A-Bot & SUCC $\uparrow$ \\ \hline
1 & GPT-2 & UpDn & 24.98 \\ 
2 & GPT-2 & LXMERT & 25.86 \\
3 & LXMERT & UpDn & 28.28 \\
4 & LXMERT & LXMERT & 33.70 \\ \hline
\end{tabular}}
\caption{The performance of the dialog system. SUCC: task success rate (\%). 
$\uparrow$: higher is better.
}
\label{dialog_system_comp_tab}
\end{table}

\begin{table*}[]
\centering
\renewcommand\arraystretch{1}
\resizebox{0.97\textwidth}{!}{
\setlength{\tabcolsep}{1.7mm}{
\begin{tabular}{l|cc|ccc|cc|cc|c} \hline
QTYPE & count-nohint & count-hint & extreme-pic & extreme-obj & extreme-obj2 & query-color & query-material  & ref-it & ref-them & all\\ \hline
UpDn & 68.89 & 75.86 & 74.12 & 66.57 & 64.47 & 84.51 & 79.00 & 86.27 & 71.77 & 73.49 \\
LXMERT & 83.23 & 87.59 & 82.92 & 81.54 & 76.75 & 91.02 & 88.54 & 88.78 & 85.63 & 85.89 \\ \hline
\end{tabular}}
}
\caption{A-Bot performance on various question subtypes. QTYPE means the question subtype.}
\label{abot_comp}
\end{table*}


\section{Dialog Generation}

\subsection{Answer Action} \label{answer_action}
The answer is divided into two types: 1) Count answer, which corresponds to the count question, gives the number of objects with specific conditions in the image. 2) Description answer responds to the extreme and refer questions, and describes one or multiple objects in natural language, e.g., `Black frame', `A decorative plate, a nightstand and a plant'. 3) Attribute answer gives the specific attribute.

\subsection{Templates} \label{all_templates}
We present all templates used for dialogue generation in Table \ref{template}.

\section{Experiments}
\subsection{Implementation Details} \label{implementation_details}
We implement our method with Pytorch and conduct all experiments on four NVIDIA Tesla V100 GPU.
For all models, we use Adam optimizer with a learning rate of 5e-5 and a mini-batch size of 32. We train QGen, A-Bot, Guesser for 10, 8, 30 epochs. For A-Bot and Guesser, we select the models with best accuracy on the val set. For QGen, we select the best performed model on the val set, under the game setting.

Formally, input sentences are tokenized by WordPiece \cite{wu2016googles} from BERT \cite{devlin2019bert}. 
We follow \citet{tan2019lxmert} to represent the visual features as a series of object representations, where objects are detected by the Faster-RCNN \cite{ren2016faster} pre-trained on Visual Genome \cite{krishna2016visual}. For each object, its representation is a concatenation of pooling features provided by \cite{anderson2018bottomup, yu2020buapt} and 4-dim vector of relative bounding box.

\subsection{Dialog System Performance Comparison} \label{dialog_system_comp}
We implement different models for this task.

\paragraph{QGen.} 1) GPT-2 \cite{radford2019language}: A decoder-only model with the pretrained language model GPT-2 as the backbone; 2) LXMERT \cite{tan2019lxmert}: Our benchmark QGen.

\paragraph{A-Bot.} 1) UpDn \cite{anderson2018bottomup}: A representative VQA model with attention mechanism; 2) LXMERT: Our benchmark A-Bot.

As shown in Figure \ref{dialog_system_comp_tab}, row 4 (QGen: LXMERT, A-Bot: LXMERT) achieves the best performance among all comparing methods, demonstrating the superiority of multimodal pretrained model.

\subsection{A-Bot Performance Comparison} \label{abot_app}
We compare UpDn \cite{anderson2018bottomup} to LXMERT \cite{tan2019lxmert} under VQA setting. As shown in Table \ref{abot_app}, LXMERT outperforms UpDn on all question subtypes, demonstrating the superiority of multimodal pretrained model.

\section{Examples}
\subsection{Image Scene Graph} \label{img_app}
We present an example of image scene graph in Figure \ref{scene_graph_case}.


\begin{table*}[]
\renewcommand\arraystretch{1}
\centering
\setlength{\tabcolsep}{7mm}{
\begin{tabular}{l|l} \hline
category & subcategories \\ \hline
home appliance & large household appliance, small household appliance \\
large household appliance & fridge, television, floor lamp, washing machine \\ 
small household appliance & coffee machine, desk lamp \\
\multirow{2}{*}{furniture} & table, chair, bench, sofa, nightstand, \\
& baby bed, cabinet, carpet, cloth tree, bed \\
table & dining table, tea table, study table \\
toy & animal toy, toy model \\
animal toy & teddy bear, elephant toy, bunny toy, giraffe toy \\
toy model & car model, airplane model, bike model, bus model \\
food & fruit, drink, meat product, baked food \\
fruit & apple, banana, watermelon \\
drink & cola, milk, tea, beer \\
baked food & bread, pizza \\
meta product & chicken leg, chicken nugget \\
sporting goods & ball, sports equipment \\
ball & soccer, basketball, tennis, bowling pin \\
sports equipment & bow, dumbbell, baseball bat, archery target, skateboard \\
kitchenware & tableware, kettle \\
tableware & plate, cup, fork, spoon \\
office supply & stationery, office equipment, paper product \\
stationery & pencil, palette \\
paper product & paperbox, notebook \\
office equipment & computer, mouse, keyboard, headphone, plug plate, phone \\ 
computer & laptop, desktop \\
decoration & vase, decorative plate, frame \\
fashion item & fashion accessory, shoes, backpack \\
fashion accessory & glasses, hat \\
shoes & boots, sandals, canvas shoes \\
hat & cotton cap, top hat, baseball cap \\
\hline 
\end{tabular}
}
\caption{The taxonomy information. The first column gives the category while the second column gives its corresponding subcategories.}
\label{table_taxonomy}
\end{table*}

\begin{table*}[]
\renewcommand\arraystretch{1}
\centering
\setlength{\tabcolsep}{5mm}{
\begin{tabular}{lll|l} 
\hline
floor &&& furniture, shoes, fridge, floor lamp, trash can, plant, table \\
dining table &&& kitchenware, drink, pizza, small household appliance, plate \\
tea table &&& decoration, book, television, cup \\
study table &&& book, office supply, toy, sporting goods \\
carpet &&& tea table, toy, sporting goods, backpack \\
plate &&& fruit, bread, meat product \\
nightstand &&& decoration, cup, glasses, hat \\
cabinet &&& decoration \\
\hline 
\end{tabular}
}
\caption{The placement relationship. The first column represents the category of objects, and the second column represents the categories that could be placed on objects of the category (in the first column).}
\label{table_placement}
\end{table*}

\begin{table*}[]
\centering
\setlength{\tabcolsep}{6mm}{
\resizebox{0.95\textwidth}{!}{

\begin{tabular}{ll|l} \hline
\multicolumn{3}{c}{\textbf{Question}} \\ \hline
\multirow{11}{*}{\large count} & \multirow{6}{*}{-nohint} & (p\_set=X) \\
& & \textit{How many [f(X)] can you see?} \\
& & \textit{How many [f(X)] are there?} \\
& & \textit{How many [f(X)] do you have?} \\
& & \textit{Can you tell me the number of [f(X)] in your image?} \\
& & \textit{I want to know the number of [f(X)] in your picture.} \\
\cline{2-3}
& \multirow{5}{*}{-hint} & (p\_set=X, count=C, copula=A) \\
& & \textit{I have [C] [f(X)], how about you?} \\
& & \textit{There [A] [C] [f(X)] in my picture, what about you?} \\
& & \textit{I can see [C] [f(X)], and you?} \\
& & \textit{[C] [f(X)] [A] in my picture, and you?} \\
\hline
\multirow{6}{*}{\large refer} & \multirow{1}{*}{-it} & \textit{What is it?} \\ 
& & \textit{Can you tell me what is it?} \\
& & \textit{Can you give me more information about it?} \\
\cline{2-3}
& \multirow{1}{*}{-them} & \textit{What are they?} \\ 
& & \textit{Can you tell me what are they?} \\
& & \textit{Can you give me more information about them?} \\
\hline
\multirow{13}{*}{\large extreme} & \multirow{5}{*}{-pic} & (p\_set=X, location=L) \\
& & \textit{What is the [L]most thing?} \\
& & \textit{What is on the far [L]?} \\
& & \textit{The [L]most on in my picture is [f(X)], what bout you?} \\
& & \textit{There is [f(X)] on the far [L] of the image, and you?} \\
\cline{2-3}
& \multirow{4}{*}{-obj} & (p\_set\_1=X1, p\_set\_2=X2, location=L) \\
& & \textit{What is the [L1]most thing on the [f(X2)]?} \\
& & \textit{The [L1]most one on the [f(X2)] is [f(X1)], what about you?} \\
& & \textit{There is [f(X1)] on the far [L] of the [f(X2)], and you?} \\
\cline{2-3}
& \multirow{4}{*}{-obj2} &(p\_set\_1=X1, p\_set\_2=X2, location\_1=L1, location\_2=L2)\\
& & \textit{What is the [L1]most thing on the [L2]most [f(X2)]?} \\
& & \textit{The [L1]smost one on the [L2]smost [f(X2)] is [f(X1)], what about you?} \\
& & \textit{There is [f(X1)] on the far [L1] of [L2]most [f(X2)], and you?} \\
\hline
\multirow{8}{*}{\large query} & \multirow{4}{*}{-color} & (p\_set=X, color=C) \\
& & \textit{What color is the [f(X)]?} \\
& & \textit{What is the color of the [f(X)]?} \\
& & \textit{The [f(X)] is [C], and you?} \\
\cline{2-3}
& \multirow{4}{*}{-material} & (p\_set=X, material=M) \\
& & \textit{What is the material of the [f(X)]?} \\
& & \textit{What is the material of the [f(X)?]} \\
& & \textit{The [f(X)] is made of [M], how about you?} \\
\hline\hline
\multicolumn{3}{c}{\textbf{Transition Sentence}} \\ 
\hline
\multicolumn{2}{c|}{\large same} & \textit{So do I / As same as you / Yes / OK.} \\
\multicolumn{2}{c|}{\multirow{2}{*}{\large diff}} & \textit{This is different from mine / mine is different from yours /} \\
& & \textit{there are some differences / it is different from mine.} \\
\multicolumn{2}{c|}{\multirow{2}{*}{\large diff}} & \textit{I have one more than you / Mine is more than yours} \\
& & \textit{Yours is less than mine.} \\
\multicolumn{2}{c|}{\large less} & \textit{Mine is less than yours / There are fewer in my image than yours.} \\
\hline
\end{tabular}
}
}
\caption{Templates used for dialog generation.}
\label{template}
\vspace{-0.4cm}
\end{table*}


\end{document}